\documentclass[conference,a4paper]{IEEEtran}

\usepackage{cite}
\usepackage{amsmath,amssymb,amsfonts}
\usepackage{algorithmic}
\usepackage{graphicx}
\usepackage{multirow}
\usepackage{textcomp}
\usepackage{xcolor}
\usepackage[ruled,vlined,linesnumbered]{algorithm2e}
\usepackage{hyperref} 

\begin{document}

\title{OMCDE: Opposition-based Multi-Stage Matrix-Based Competitive Differential Evolution}

\author{
  \IEEEauthorblockN{Mohammad Sarhangzadeh}
  \IEEEauthorblockA{Dept. of Computer Engineering \\
    Bilkent University \\
    Ankara, Turkey \\
    m.sarhangzadeh@bilkent.edu.tr}
  \and
  \IEEEauthorblockN{Shahryar Rahnamayan}
  \IEEEauthorblockA{Department of Engineering \\
    Brock University \\
    St. Catharines, Canada \\
    Srahnamayan@brocku.ca}
  \and
  \IEEEauthorblockN{Doruk Oner}
  \IEEEauthorblockA{Dept. of Computer Engineering \\
    Bilkent University \\
    Ankara, Turkey \\
    doruk.oner@bilkent.edu.tr}
}

\maketitle

\noindent{\footnotesize\textcopyright{} 2026 IEEE. Personal use of this material is permitted. Permission from IEEE must be obtained for all other uses, in any current or future media, including reprinting/republishing this material for advertising or promotional purposes, creating new collective works, for resale or redistribution to servers or lists, or reuse of any copyrighted component of this work in other works. The published version is available at \url{https://ieeexplore.ieee.org/document/11727162}.\par}\medskip

\begin{abstract}
Differential Evolution has proven to be an effective algorithm for continuous global optimization, but it's modern high-performing variants often introduce additional complexity and computational cost. We propose OMCDE, a matrix-based competitive Differential Evolution algorithm that aims to improve optimization performance while maintaining computational efficiency. The proposed approach integrates competition-driven search dynamics, adaptive behavior between optimization stages, and opposition-based initialization within a parallelizable framework. Empirical results on widely used CEC 2017 functions show that OMCDE achieves strong performance compared to several state-of-the-art DE algorithms, while remaining efficient and scalable. These findings indicate that OMCDE is a promising alternative for solving complex numerical optimization problems.
\end{abstract}


\section{Introduction}
Differential Evolution (DE) \cite{storn1997differential} is a powerful population-based stochastic optimizer known for its simplicity and effectiveness in solving global numerical optimization problems. A fundamental challenge in its design is balancing exploration and exploitation. Exploration refers to the algorithm’s ability to search broad, unknown regions of the solution space to maintain diversity and avoid local optima. Conversely, exploitation focuses on refining the search within promising local neighborhoods to accelerate convergence toward the global optimum.

The development of L-SHADE \cite{tanabe2014improving} marked a significant shift in modern DE by introducing success-history-based parameter adaptation and linear population size reduction. Building on this, recent research increasingly focuses on utilizing individual-level evolutionary information rather than relying solely on population-wide settings \cite{ahmad2022differential,pant2020differential}. This information typically includes the specific fitness of an individual, its historical success in producing better offspring, and its relative standing within the current population. This shift allows algorithms to tailor search strategies to the unique landscape each candidate individual is currently navigating, adjusting parameters like the scaling factor ($F$) and crossover rate ($CR$) for every member of the population.

While these individual-level strategies improve search quality, they often come with a trade-off in speed. Many current methods require complex sorting, historical record-keeping, and sequential loops that make them computationally slow and difficult to scale, especially for high-dimensional problems. Furthermore, these methods often rely on fixed logic that may not adapt well as the search environment and population diversity change over time.

To address these challenges, we propose OMCDE (Opposition-based Matrix-Based Competitive Differential Evolution). Our approach adopts the triplet-competition mechanism introduced by TCDE \cite{yang2024triple}, which mirrors the natural process where individuals within a population compete for survival. In this model, candidates are ranked within small groups and assigned mutation roles Best, Medium, or Worst to balance exploration and local refinement based on their competitive standing.

We further extend this by incorporating a stage-dependent mutation strategy, inspired by how biological organisms adapt not just to each other, but to the shifting pressures of their environment. In nature, survival strategies often become more aggressive and focused as resources reach their limit; similarly, OMCDE transitions from broad, diversified exploration to intense, greedy exploitation in the final stages of the search to refine the global optimum. By integrating this competitive logic with opposition-based initialization and a fully matrix-based framework, we enable parallel execution that significantly reduces runtime by eliminating computationally expensive nested loops.

We evaluated OMCDE on the CEC 2017 \cite{wu2016problem} benchmark suite, where it achieved performance levels near current state-of-the-art variants while remaining exceptionally fast\footnote{\url{https://github.com/MSarhangzadeh/OMCDE-Optimization}}. The main contributions of this study are as follows:

\begin{enumerate} \item A unified matrix-based architecture that reformulates core DE operators into vectorized computations, enabling simultaneous population-wide updates and high-speed execution. \item A stage-dependent mutation scheme that adapts to the search progress, transitioning from diversified exploration to focused, greedy exploitation. \item A framework utilizing triplet-based competition managing exploration and exploitation at the individual level.
\end{enumerate}

\section{Related Works}

While the fundamental operators and standard evolutionary cycle of Differential Evolution (DE) are well-established in the literature \cite{pant2020differential,ahmad2022differential}, this section focuses on the high-level paradigm shifts that define current state-of-the-art variants.

Modern research in DE was advanced by JADE \cite{zhang2009jade}, which introduced the DE/current-to-pbest/1 mutation strategy:
\begin{equation}
v_{i,g} = x_{i,g} + F \cdot (x_{pbest,g} - x_{i,g}) + F \cdot (x_{r1,g} - x_{r2,g})
\end{equation}
where $x_{pbest,g}$ is randomly selected from the top $p\%$ of the population. This scheme balances exploration and exploitation by using the vector $(x_{pbest,g} - x_{i,g})$ to pull the individual toward the best-known regions, while $(x_{r1,g} - x_{r2,g})$ maintains the stochastic search power needed to discover new areas. This foundation was solidified by L-SHADE \cite{tanabe2014improving}, which added Linear Population Size Reduction to dynamically shrink the population as the search progresses. By reducing the number of individuals, L-SHADE effectively shifts the computational budget from broad exploration in early stages to intensive local refinement toward the end of the search.

Building on the success of L-SHADE, various descendants emerged to further exploit individual-level evolutionary information. Some, like iLSHADE-RSP \cite{choi2021improved}, introduced stochastic perturbations using Cauchy distributions, leveraging "heavier tails" to enable large steps that help the population escape local optima. Others, most notably LSHADE-RSP \cite{stanovov2018lshade}, refined the selection process with Rank-Based Selective Pressure (RSP). By assigning higher selection probabilities to better-ranked individuals rather than treating them equally, RSP increases the greediness of the mutation operator, significantly accelerating convergence on complex landscapes.

The shift toward treating individuals differently based on their standing led to TCDE \cite{yang2024triple}. Inspired by natural competition, TCDE groups the population into random triplets where individuals are ranked as $x_{best}$, $x_{medium}$, and $x_{worst}$. Leveraging the adaptive parameter logic of L-SHADE, TCDE assigns specific control parameters ($F$ and $CR$) to each member based on their role. These roles dictate the mutation strategy, ensuring a balance between local refinement and global exploration:

\begin{equation}
\label{eq:tcde_best}
    v_{best} = x_{best} + F_{best} \cdot (x_{r1} - x_{r2})
\end{equation}

\begin{equation}
\label{eq:tcde_med}
\begin{split}
    v_{medium} = & \, x_{medium} + F_{med} \cdot (x_{best} - x_{medium}) \\
                 & + F_{med} \cdot (x_{r1} - x_{r2})
\end{split}
\end{equation}

\begin{equation}
\label{eq:tcde_worst}
\begin{split}
    v_{worst} = & \, x_{worst} + F_{worst} \cdot (x_{best} - x_{worst}) \\
                & + F_{worst} \cdot (x_{med} - x_{worst})
\end{split}
\end{equation}

where $\mathbf{x}_{best}, \mathbf{x}_{med}, \mathbf{x}_{worst}$ are the sorted individuals within the triplet, and $\mathbf{x}_{r1}, \mathbf{x}_{r2}$ are random vectors sampled from the union of the population and the external archive $\mathbf{A}$. The parameter $F$ denotes the adaptive scaling factor associated with each individual.

In this hierarchical structure, the mutation strategies are tailored to the individual's rank. The best individuals focus on exploiting their local neighborhood to refine the solution, while the worst individuals are driven to explore new regions of the search space to maintain diversity.

More recently, the RDE algorithm \cite{tao2024efficient} demonstrated that recombining successful strategies can further push performance limits. RDE integrates two distinct mutation operators: the classic current-to-pbest/1 \cite{zhang2009jade} and the sorted current-to-order-pbest/1 \cite{yang2024triple}. Unlike static ensembles, RDE employs an adaptive resource allocation mechanism that tracks the cumulative fitness improvement generated by each strategy. It then dynamically adjusts the population ratio assigned to each operator, ensuring that computational resources naturally shift toward the strategy that is currently navigating the landscape most effectively.

While TCDE and RDE effectively utilize competitive relationships, their reliance on iterative sorting and complex history management creates a significant computational bottleneck in high dimensions. OMCDE overcomes these limitations by embedding competitive dynamics into a high-throughput matrix-based framework and introducing a stage-dependent strategy that explicitly shifts the algorithm's structural behavior from exploration to greedy exploitation.

\section{Matrix-Based Operations}

\subsection{Motivation}
Traditional DE \cite{storn1997differential} implementations rely on iterative loops to process individuals sequentially, a design that inherently fails to exploit the parallel architecture of modern CPUs and GPUs. While the concept of vectorized DE was popularized in early frameworks\cite{price2005differential}, and recent research has proposed matrix-based structures for vanilla algorithms \cite{li2024matrix, zhan2021matrix}, these approaches have mostly been limited to standard vanilla algorithms. In this work, we extend this matrix-based paradigm to OMCDE by reformulating the entire evolutionary cycle: mutation, crossover, selection, and even dynamic archive management into vectorized linear algebra operations. This unified framework allows for the simultaneous update of the entire population, significantly reducing computational overhead and enabling the use of high-performance computing resources without the need for complex low-level programming.
\subsection{Matrix-Based Operations}

Let $N$ denote the population size and $D$ the dimensionality of the search space. Instead of treating individuals as separate objects, we represent the entire population as a matrix $\mathbf{X}$ with dimensions $N \times D$, where each row corresponds to a candidate solution and each column represents a specific dimension. We similarly define the boundary vectors $\mathbf{L}$ and $\mathbf{U}$ with dimensions of $1 \times D$ representing the lower and upper bounds of the search space.

To facilitate vectorized operations, we utilize two auxiliary primitives: a column vector of ones, $\mathbf{1}_{N \times 1}$, used to replicate constraints across the population axis; and a random matrix, $\mathbf{R}_{N \times D}$, containing uniformly distributed random numbers in $[0, 1]$.

The initialization of the population is performed via a single matrix operation. We use the Hadamard product to broadcast the boundary constraints across the random matrix, projecting values into the feasible space immediately:
\begin{equation}
\begin{split}
\mathbf{X}_{N \times D} = & \, (\mathbf{1}_{N \times 1} \times \mathbf{L}_{1 \times D}) \\
& + (\mathbf{1}_{N \times 1} \times (\mathbf{U}_{1 \times D} - \mathbf{L}_{1 \times D})) \circ \mathbf{R}_{N \times D}
\end{split}
\end{equation}

Following initialization and at every subsequent generation, the fitness of the entire population is evaluated in a single parallel step, mapping the objective function across the matrix rows to produce a fitness vector:
\begin{equation}
\mathbf{Fit}_{N \times 1} = f(\mathbf{X}_{N \times D})
\end{equation}

In this framework, conditional logic is strictly replaced by Boolean Mask Matrices to avoid branch divergence. For mutation, we row-shuffle the population indices to create permuted matrices $\mathbf{X}_{r1}$, $\mathbf{X}_{r2}$, and $\mathbf{X}_{r3}$. The difference vector calculation is performed as a single block, where the mutation factor $\mathbf{F}$ is defined as a column vector ($N \times 1$) to allow unique adaptive parameters for every individual:
\begin{equation}
\begin{split}
\mathbf{V}_{N \times D} = & \, \mathbf{X}_{r1, N \times D} \\
& + \mathbf{F}_{N \times 1} \circ (\mathbf{X}_{r2, N \times D} - \mathbf{X}_{r3, N \times D})
\end{split}
\end{equation}

Crossover is similarly vectorized by generating a random matrix and comparing it against the crossover rate vector $\mathbf{CR}_{N \times 1}$ to generate a binary mask $\mathbf{M}_{N \times D}$, where $1$ indicates a crossover event. The trial population $\mathbf{T}$ is constructed by blending the mutant matrix $\mathbf{V}$ and the target matrix $\mathbf{X}$ using the mask:
\begin{equation}
\label{eq:matrix_cross}
\begin{split}
\mathbf{T}_{N \times D} = & \, \mathbf{M}_{N \times D} \circ \mathbf{V}_{N \times D} \\
& + (\mathbf{1}_{N \times 1} - \mathbf{M}_{N \times D}) \circ \mathbf{X}_{N \times D}
\end{split}
\end{equation}

Selection is performed by evaluating the fitness of the trial matrix $\mathbf{T}$ and comparing it against $\mathbf{X}$ to generate a selection mask $\mathbf{S}_{N \times 1}$. The population update is a masked assignment:
\begin{equation}
\label{eq:matrix_select}
\begin{split}
\mathbf{X}_{new, N \times D} = & \, \mathbf{S}_{N \times 1} \circ \mathbf{T}_{N \times D} \\
& + (\mathbf{1}_{N \times 1} - \mathbf{S}_{N \times 1}) \circ \mathbf{X}_{old, N \times D}
\end{split}
\end{equation}

Finally, we vectorize the management of historical archives for parameter adaptation. Instead of loop-based appending, we utilize boolean indexing to extract successful parameters in batch using the selection mask $\mathbf{S}$. The archives are updated using vectorized dot products between fitness improvement vectors and successful parameter vectors. This formulation allows the adaptive logic to scale efficiently with population size, avoiding sequential processing bottlenecks.

\section{Proposed OMCDE Algorithm}
We propose the Opposition-based Matrix-Based Competitive Differential Evolution (OMCDE). Our main goal is to balance exploration and exploitation while making the algorithm run extremely fast. To do this, OMCDE combines four key mechanisms into a unified matrix framework: Opposition-Based Initialization \cite{rahnamayan2008opposition}, Triplet Competition \cite{yang2024triple}, Two-Stage Heterogeneous Mutation, and Adaptive Parameter Control with Linear Population Size Reduction \cite{brest2016shade}. The algorithm is smart; it uses the evolutionary information of each individual to assign specific strategies based on their performance rank. It also uses a stage-dependent mutation strategy, inspired by the environment's effect on evolution, that dynamically adapts the mutation behavior according to the search progress. Finally, because every logic step is converted into vectorized matrix operations, the algorithm is exceptionally fast, maintaining high performance even in high-dimensional search spaces.

\subsection{Opposition-Based Matrix Initialization}
Standard uniform initialization relies on stochasticity, often placing the population in all regions. To accelerate convergence, we integrate Opposition-Based Learning. However, unlike traditional methods \cite{rahnamayan2008opposition} that require sorting and slicing a combined population of $2N$ individuals which has an overhead of $O(N \log N)$, we employ a high-speed Population-Level Selection mechanism based on total fitness.

First, we generate the opposition matrix $\mathbf{X}_{opp}$ in a single broadcasted operation using the boundary vectors $\mathbf{L}$ and $\mathbf{U}$:

\begin{equation}
\label{eq:opp_init}
\mathbf{X}_{opp, N \times D} = (\mathbf{1}_{N \times 1} \times (\mathbf{L} + \mathbf{U})) - \mathbf{X}_{N \times D}
\end{equation}

Both the original population $\mathbf{X}$ and the opposition population $\mathbf{X}_{opp}$ are evaluated in parallel to produce fitness vectors $\mathbf{Fit}$ and $\mathbf{Fit}_{opp}$. To maintain maximum computational throughput, we simply compare the total fitness of both matrices:

\begin{equation}
\label{eq:opp_select}
\mathbf{X}^{(0)} = 
\begin{cases} 
\mathbf{X}_{opp} & \text{if } \sum \mathbf{Fit}_{opp} < \sum \mathbf{Fit} \\
\mathbf{X} & \text{otherwise}
\end{cases}
\end{equation}

This approach allows us to instantaneously select the statistically superior population distribution without breaking the vectorized workflow, ensuring the initialization phase remains $O(N)$ and exceptionally fast.

\subsection{Triplet Competition Mechanism}
OMCDE employs a hierarchical structure where individuals compete within random triplets rather than the entire population, preventing the premature convergence typical of global selection strategies. To avoid the computational overhead of iterative sorting, we utilize tensor reshaping. The population matrix $\mathbf{X}$ ($N \times D$) is instantly reshaped into a tensor of $N/3$ triplets and sorted in parallel along the triplet axis. This operation efficiently stratifies the population into three sub-matrices $\mathbf{X}_{best}$, $\mathbf{X}_{med}$, and $\mathbf{X}_{worst}$, where the best group contains individuals with the lowest fitness values designated for exploitation, and the worst group consists of those with the highest fitness values targeted for exploration. This enables the simultaneous application of heterogeneous mutation strategies without the need for complex individual ranking.

\subsection{Stage-Dependent Heterogeneous Mutation}
To effectively balance global exploration with local exploitation, OMCDE employs a dynamic mutation strategy. The algorithm switches between two distinct behavioral modes based on the $FES / FES_{max}$ ratio.

In the first phase $FES \le \lambda \cdot FES_{max}$, where $\lambda$ represents the stage switching parameter that defines the duration of the initial diversity maintenance period, preserving diversity is critical to prevent premature convergence. We employ the standard competitive strategy established in TCDE \cite{yang2024triple}, where the mutation logic is strictly dictated by the individual's rank within its triplet. The superior individuals ($\mathbf{X}_{best}$) exploit their local neighborhood (Eq. \ref{eq:tcde_best}), the intermediate individuals ($\mathbf{X}_{med}$) employ a balanced strategy (Eq. \ref{eq:tcde_med}), and the inferior individuals ($\mathbf{X}_{worst}$) are forced to explore by learning from the superior vectors to escape poor regions (Eq. \ref{eq:tcde_worst}). Note that to prevent path duplication, the random difference vectors are selected from the union of the current population and the external Solution Archive $\mathbf{A}$ implemented based on \cite{zhang2009jade}. This archive preserves diversity by storing parent vectors recently replaced by superior offspring, allowing the algorithm to leverage historical genetic information to escape local optima.

As the search nears completion ($FES >  \lambda \cdot FES_{max}$), the algorithm shifts to a greedy convergence mode. In this phase, the priority changes from diversity maintenance to rapid refinement of the best known solution. We first construct a global best matrix $\mathbf{G}_{best}$ by broadcasting the single global best vector $\mathbf{x}_{gbest}$ to match the dimensions of the sub-populations. All three groups then shift their strategies to pull strongly toward this global optimum while retaining their hierarchical learning structure. The Best individuals are pulled toward the Global Best while retaining a difference vector for local fine-tuning:
\begin{equation}
\label{eq:mut_late_best}
\begin{split}
\mathbf{V}_{best, \frac{N}{3} \times D} = & \, \mathbf{X}_{best} + \mathbf{F}_{best} \circ (\mathbf{G}_{best} - \mathbf{X}_{best}) \\
& + \mathbf{F}_{best} \circ (\mathbf{X}_{r1} - \mathbf{X}_{r2})
\end{split}
\end{equation}

The Medium individuals are pulled toward the Global Best, but also continue to learn from the Local Best to bridge the search space:

\begin{equation}
\label{eq:mut_late_med}
\begin{split}
\mathbf{V}_{med, \frac{N}{3} \times D} = & \, \mathbf{X}_{med} + \mathbf{F}_{med} \circ (\mathbf{G}_{best} - \mathbf{X}_{med}) \\
& + \mathbf{F}_{med} \circ (\mathbf{X}_{best} - \mathbf{X}_{med}) \\
& + \mathbf{F}_{med} \circ (\mathbf{X}_{r1} - \mathbf{X}_{r2})
\end{split}
\end{equation}

Finally, the Worst individuals use the most aggressive strategy, being pulled toward the Global Best, the Local Best, and the Local Medium simultaneously to rapidly close the gap:

\begin{equation}
\label{eq:mut_late_worst}
\begin{split}
\mathbf{V}_{worst, \frac{N}{3} \times D} = & \, \mathbf{X}_{worst} + \mathbf{F}_{worst} \circ (\mathbf{G}_{best} - \mathbf{X}_{worst}) \\
& + \mathbf{F}_{worst} \circ (\mathbf{X}_{best} - \mathbf{X}_{worst}) \\
& + \mathbf{F}_{worst} \circ (\mathbf{X}_{med} - \mathbf{X}_{worst})
\end{split}
\end{equation}
The implementation of this multi-stage mechanism is driven by the observation that optimal search behavior changes over time. By enforcing exploration in the early stages and switching to greedy exploitation in the final quartile, OMCDE mimics the natural adaptation of organisms to shrinking resources, ensuring that the population explores the landscape broadly before focusing efficiently onto the global optimum.

\begin{algorithm}[t]
\SetAlgoLined
\DontPrintSemicolon
\caption{Opposition-Controlled Matrix-Based Competitive Differential Evolution (OMCDE)}
\label{alg:omcde}

\KwIn{$N_{init}$, $D$, $FES_{max}$, Bounds $[\mathbf{L}, \mathbf{U}]$}
\KwOut{Global Best Solution $\mathbf{x}_{gbest}$}

\textbf{1. Initialization:}\;
Generate random population $\mathbf{X}$ and opposition matrix $\mathbf{X}_{opp}$ by Eq. \ref{eq:opp_init}\;
Select initial $\mathbf{X}^{(0)}$ by comparing total fitness sums by Eq. \ref{eq:opp_select}\;
Initialize history archives $\mathbf{M}_F, \mathbf{M}_{CR}$ with $0.5$, set $FES = N_{init}$\;

\While{$FES < FES_{max}$}{
    \textbf{2. Linear Reduction \& Parameter Generation:}\;
    Calculate new size $N_{next}$ by Eq. \ref{eq:lpsr} and truncate $\mathbf{X}$ if $N_{next} < N$\;
    Generate $\mathbf{F}, \mathbf{CR}$ vectors utilizing Cauchy/Normal distributions by Eq. \ref{eq:param_gen}\;
    
    \textbf{3. Triplet Competition (Matrix Reshaping):}\;
    Reshape $\mathbf{X}$ into tensor $(N/3 \times 3 \times D)$ and sort parallel along axis 1\;
    Partition into sub-matrices $\mathbf{X}_{best}, \mathbf{X}_{med}, \mathbf{X}_{worst}$\;
    
    \textbf{4. Stage-Dependent Mutation:}\;
    \eIf{$FES \le \lambda \cdot FES_{max}$}{
        \tcp{Phase 1: Competitive Exploration}
        Compute $\mathbf{V}_{best}, \mathbf{V}_{med}, \mathbf{V}_{worst}$ utilizing TCDE rules (Eq. \ref{eq:tcde_best}--\ref{eq:tcde_worst})\;
    }{
        \tcp{Phase 2: Greedy Exploitation}
        Broadcast $\mathbf{x}_{gbest}$ to matrix $\mathbf{G}_{best}$\;
        Compute $\mathbf{V}_{best}, \mathbf{V}_{med}, \mathbf{V}_{worst}$ utilizing Greedy rules (Eq. \ref{eq:mut_late_best}--\ref{eq:mut_late_worst})\;
    }
    
    \textbf{5. Vectorized Crossover \& Selection:}\;
    Concatenate sub-matrices to form global mutant $\mathbf{V}$\;
    Generate trial matrix $\mathbf{T}$ by Eq. \ref{eq:matrix_cross}\;
    Evaluate fitness $\mathbf{Fit}_{trial}$ and create selection mask $\mathbf{S}$\;
    Update population $\mathbf{X}$ by Eq. \ref{eq:matrix_select}\;
    
    \textbf{6. Adaptation:}\;
    Archive rejected solutions into $\mathbf{A}$\;
    Update $\mathbf{M}_F, \mathbf{M}_{CR}$ utilizing Weighted Lehmer Mean by Eq. \ref{eq:lehmer_mean}\;
    
    $FES = FES + N$\;
}
\end{algorithm}

\subsection{Crossover and Selection}
Following the heterogeneous mutation stage, the three mutant sub populations are vertically concatenated to reconstruct the global mutant matrix $\mathbf{V}$ with dimensions $N \times D$. We then employ a binomial crossover strategy adapted for matrix execution. A random matrix is compared against the crossover rate vector $\mathbf{CR}$ to create a binary mask, while a secondary mask enforces the guaranteed inheritance condition. The final trial population $\mathbf{T}$ is constructed by blending the mutant and target matrices using the masked operation defined in Eq. \ref{eq:matrix_cross}.

Selection is performed via a parallel greedy comparison between the trial matrix $\mathbf{T}$ and the target matrix $\mathbf{X}$. The fitness of the trial population is evaluated in a single batch step, and a boolean selection mask $\mathbf{S}_{N \times 1}$ is generated to identify trial vectors that outperform their targets. The population update is then executed instantaneously using the masking operation in Eq. \ref{eq:matrix_select}, discarding inferior solutions while retaining the superior ones for the next generation.

\subsection{Adaptive Parameter Control}
The performance of differential evolution is highly sensitive to the control parameters $F$ and $CR$. To avoid manual tuning, OMCDE utilizes a history based adaptation mechanism \cite{tanabe2013success}. However, unlike standard loop based implementations that generate parameters sequentially, we perform these operations as fully vectorized batch processes.

At the beginning of each generation, we generate parameter vectors $\mathbf{F}$ and $\mathbf{CR}$ for the entire population simultaneously. We maintain two history memory archives, $\mathbf{M}_F$ and $\mathbf{M}_{CR}$, of size $H$. A vector of random indices $\mathbf{r}$ is generated to sample these archives. The parameters are then constructed by broadcasting the sampled memory values and adding noise derived from Cauchy and Normal distributions, respectively:

\begin{equation}
\label{eq:param_gen}
\begin{split}
\mathbf{F}_{N \times 1} &= \text{Cauchy}(\mathbf{M}_{F}[\mathbf{r}], 0.1), \\
\mathbf{CR}_{N \times 1} &= \mathcal{N}(\mathbf{M}_{CR}[\mathbf{r}], 0.1)
\end{split}
\end{equation}

After the selection step, the memory archives are updated to reflect the most successful strategies \cite{tanabe2013success}. We first identify the subset of successful parameters ($\mathbf{S}_F, \mathbf{S}_{CR}$) corresponding to trial vectors that improved fitness. We calculate the improvement weights $\Delta \mathbf{f}$ as the absolute difference between the trial and target fitness values. The update value is computed using the Weighted Lehmer Mean. In our matrix framework, this complex summation is reduced to a high-speed vectorized dot product between the weight vector and the parameter powers:

\begin{equation}
\label{eq:lehmer_mean}
mean_{WL} = \frac{\Delta \mathbf{f}^T_{1 \times S} \cdot (\mathbf{S}_{F, S \times 1})^{\circ 2}}{\Delta \mathbf{f}^T_{1 \times S} \cdot \mathbf{S}_{F, S \times 1}}
\end{equation}
where $\mathbf{S}_{F}$ is the vector containing the scaling factors of successful individuals, and $\Delta \mathbf{f}$ represents their corresponding fitness improvements. This computes a weighted mean where parameters resulting in larger fitness gains contribute more significantly to the history update.

Finally, the history memory at position $k$ is updated using a linear combination of the old value and the new Lehmer mean:
\begin{equation}
\label{eq:mem_update}
\mathbf{M}_{F}[k] = 0.5 \cdot \mathbf{M}_{F}[k] + 0.5 \cdot mean_{WL}
\end{equation}
This vectorized formulation allows OMCDE to adapt its search behavior dynamically based on population-wide feedback without incurring the computational penalty of iterative history processing.

To further enhance computational efficiency and accelerate convergence, we dynamically reduce the population size $N$ as the search progresses. This strategy mimics natural resource scarcity, forcing the algorithm to focus its computational power on the most promising individuals during the critical later stages. The population size for the next generation is calculated as:
\begin{equation}
\label{eq:lpsr}
N_{g+1} = \text{round} \left[ \left( \frac{N_{min} - N_{init}}{FES_{max}} \right) \cdot FES + N_{init} \right]
\end{equation}
where $N_{init}$ and $N_{min}$ represent the initial and minimum population sizes, and $FES_{max}$ denotes the maximum number of function evaluations. This mechanism linearly decreases the population size as the search progresses, allocating more computational resources to the best individuals in later stages.

If $N_{g+1} < N_{curr}$, we sort the population matrix $\mathbf{X}$ by fitness and truncate the worst individuals. This reduction creates a naturally increasing selective pressure and ensures the memory footprint of the algorithm scales down linearly.

The complete computational flow of the proposed method is summarized in Algorithm \ref{alg:omcde}. It is crucial to highlight that, unlike traditional implementations that rely on nested loops to process individuals sequentially, Algorithm \ref{alg:omcde} contains only a single outer loop governing the generations. All internal operations, including mutation, crossover, and selection, are executed as simultaneous matrix instructions, ensuring maximum computational efficiency.

\section{Experiments}
\begin{table*}[htbp]
\caption{Comparative Results of OMCDE vs. Competitors on CEC 2017 ($D=30$) over 31 Independent Runs. Symbols $+$/$=$/$-$ indicate OMCDE is significantly better than / similar to / worse than the competitor (Wilcoxon rank-sum test, $\alpha=0.05$).}
\label{tab:cec2017}
\centering

\resizebox{\textwidth}{!}{%
\begin{tabular}{lcccccccccccc}
\hline
\multirow{2}{*}{\textbf{Func}} & \multicolumn{2}{c}{\textbf{OMCDE}} & \multicolumn{2}{c}{\textbf{TCDE}} & \multicolumn{2}{c}{\textbf{LSHADE-RSP}} & \multicolumn{2}{c}{\textbf{iLSHADE-RSP}} & \multicolumn{2}{c}{\textbf{L-SHADE}} & \multicolumn{2}{c}{\textbf{JADE}} \\
\cline{2-13}
 & Mean & SD & Mean & SD & Mean & SD & Mean & SD & Mean & SD & Mean & SD \\
\hline
\multicolumn{13}{c}{\textit{Unimodal Functions}} \\
$F_1$ & \textbf{0.00E+00} & \textbf{0.00E+00} & 0.00E+00$=$ & 0.00E+00 & 0.00E+00$=$ & 0.00E+00 & 0.00E+00$=$ & 0.00E+00 & 0.00E+00$=$ & 0.00E+00 & 0.00E+00$=$ & 0.00E+00 \\
$F_3$ & \textbf{0.00E+00} & \textbf{0.00E+00} & 0.00E+00$=$ & 0.00E+00 & 0.00E+00$=$ & 0.00E+00 & 0.00E+00$=$ & 0.00E+00 & 0.00E+00$=$ & 0.00E+00 & 9.43E+03$+$ & 1.54E+04 \\
\hline
\multicolumn{13}{c}{\textit{Simple Multimodal Functions}} \\
$F_4$ & 5.86E+01 & 0.00E+00 & 5.86E+01$=$ & 0.00E+00 & 5.86E+01$=$ & 0.00E+00 & 2.27E+01$-$ & 7.02E-01 & 5.87E+01$=$ & 7.70E-01 & 4.88E+01$-$ & 2.39E+01 \\
$F_5$ & \textbf{3.52E+00} & 1.62E+00 & 4.83E+00$+$ & 1.31E+00 & 7.11E+00$+$ & 2.09E+00 & 7.89E+00$+$ & 2.38E+00 & 6.77E+00$+$ & 1.60E+00 & 2.63E+01$+$ & 4.00E+00 \\
$F_6$ & 0.00E+00 & 2.53E-08 & 8.54E-06$+$ & 5.44E-06 & 0.00E+00$=$ & 2.71E-08 & 5.77E-08$+$ & 1.34E-07 & 0.00E+00$=$ & 1.92E-08 & 0.00E+00$=$ & 0.00E+00 \\
$F_7$ & \textbf{3.60E+01} & 1.11E+00 & 3.64E+01$+$ & 9.10E-01 & 3.95E+01$+$ & 2.50E+00 & 4.08E+01$+$ & 3.42E+00 & 3.77E+01$+$ & 1.42E+00 & 5.46E+01$+$ & 4.02E+00 \\
$F_8$ & \textbf{5.26E+00} & 1.48E+00 & 6.92E+00$+$ & 1.84E+00 & 7.38E+00$+$ & 2.28E+00 & 7.93E+00$+$ & 2.39E+00 & 7.24E+00$+$ & 1.59E+00 & 2.64E+01$+$ & 3.83E+00 \\
$F_9$ & 0.00E+00 & 0.00E+00 & 0.00E+00$=$ & 0.00E+00 & 0.00E+00$=$ & 0.00E+00 & 0.00E+00$=$ & 0.00E+00 & 0.00E+00$=$ & 0.00E+00 & 3.51E-03$=$ & 1.75E-02 \\
$F_{10}$ & \textbf{1.46E+03} & 3.06E+02 & 1.98E+03$+$ & 1.88E+02 & 1.92E+03$+$ & 3.31E+02 & 1.90E+03$+$ & 3.46E+02 & 1.49E+03$=$ & 1.51E+02 & 1.92E+03$+$ & 2.15E+02 \\
\hline
\multicolumn{13}{c}{\textit{Hybrid Functions}} \\
$F_{11}$ & 1.91E+01 & 2.42E+01 & 2.01E+01$=$ & 2.57E+01 & 2.62E+00$-$ & 2.23E+00 & 3.41E+00$-$ & 5.57E+00 & 28.8E+01$+$ & 2.81E+01 & 3.68E+01$+$ & 2.65E+01 \\
$F_{12}$ & 1.84E+02 & 1.29E+02 & 2.27E+02$=$ & 2.36E+02 & 9.51E+01$-$ & 7.16E+01 & 1.20E+02$=$ & 7.85E+01 & 1.06E+09$+$ & 3.76E+02 & 1.15E+03$+$ & 3.87E+02 \\
$F_{13}$ & \textbf{1.09E+01} & 9.66E+00 & 1.58E+01$+$ & 9.64E+00 & 1.73E+01$+$ & 5.41E+00 & 1.80E+01$+$ & 4.92E+00 & 1.72E+01$+$ & 4.75E+00 & 4.79E+01$+$ & 5.92E+01 \\
$F_{14}$ & \textbf{2.10E+01} & 4.13E+00 & 2.21E+01$+$ & 1.27E+00 & 2.15E+01$=$ & 1.23E+00 & 2.17E+01$+$ & 1.06E+00 & 2.16E+01$+$ & 1.24E+00 & 7.30E+03$=$ & 2.70E+04 \\
$F_{15}$ & \textbf{1.11E+00} & 1.85E-01 & 1.73E+00$+$ & 5.86E-01 & 1.34E+00$=$ & 7.73E-01 & 1.18E+00$=$ & 7.50E-01 & 3.10E+00$+$ & 1.46E+00 & 7.71E+02$+$ & 1.82E+03 \\
$F_{16}$ & 2.53E+01 & 1.06E+01 & 3.69E+01$=$ & 4.12E+01 & 2.87E+01$=$ & 4.35E+01 & 1.86E+01$-$ & 6.87E+00 & 6.25E+01$+$ & 7.43E+01 & 3.83E+02$+$ & 1.45E+02 \\
$F_{17}$ & 4.02E+01 & 9.45E+00 & 4.48E+01$+$ & 1.15E+01 & 3.78E+01$=$ & 7.04E+00 & 3.89E+01$=$ & 7.18E+00 & 3.31E+01$-$ & 6.94E+00 & 7.70E+01$+$ & 3.12E+01 \\
$F_{18}$ & 2.10E+01 & 3.42E-01 & 2.14E+01$+$ & 5.04E-01 & 2.08E+01$-$ & 2.89E-01 & 2.08E+01$-$ & 2.88E-01 & 2.19E+01$+$ & 1.07E+00 & 1.58E+04$=$ & 4.77E+04 \\
$F_{19}$ & \textbf{3.46E+00} & 1.34E+00 & 4.18E+00$+$ & 1.78E+00 & 3.49E+00$=$ & 1.08E+00 & 3.49E+00$=$ & 6.06E-01 & 5.38E+00$+$ & 1.40E+00 & 1.27E+03$=$ & 3.62E+03 \\
$F_{20}$ & \textbf{3.21E+01} & 1.00E+01 & 3.86E+01$+$ & 9.20E+00 & 3.38E+01$=$ & 9.50E+00 & 3.24E+01$=$ & 7.02E+00 & 4.10E+01$+$ & 8.81E+00 & 1.21E+02$+$ & 6.10E+01 \\
\hline
\multicolumn{13}{c}{\textit{Composition Functions}} \\
$F_{21}$ & \textbf{2.04E+02} & 1.02E+00 & 2.22E+02$+$ & 1.87E+00 & 2.07E+02$+$ & 2.19E+00 & 2.08E+02$+$ & 2.36E+00 & 2.07E+02$+$ & 1.49E+00 & 2.27E+02$+$ & 5.36E+00 \\
$F_{22}$ & \textbf{1.00E+02} & 0.00E+00 & 1.10E+02$+$ & 0.00E+00 & 1.00E+02$=$ & 0.00E+00 & 1.00E+02$=$ & 0.00E+00 & 1.00E+02$=$ & 0.00E+00 & 1.45E+02$=$ & 3.18E+02 \\
$F_{23}$ & 3.52E+02 & 2.86E+00 & 3.52E+02$=$ & 3.07E+00 & 3.51E+02$=$ & 3.47E+00 & 3.50E+02$-$ & 3.29E+00 & 3.49E+02$-$ & 2.70E+00 & 3.74E+02$+$ & 6.00E+00 \\
$F_{24}$ & \textbf{4.20E+02} & 1.75E+00 & 4.32E+02$+$ & 2.18E+00 & 4.27E+02$+$ & 2.10E+00 & 4.26E+02$+$ & 2.45E+00 & 4.26E+02$+$ & 1.67E+00 & 4.40E+02$+$ & 4.50E+00 \\
$F_{25}$ & 3.87E+02 & 1.73E-02 & 3.87E+02$=$ & 1.54E-02 & 3.87E+02$=$ & 0.00E+00 & 3.79E+02$-$ & 0.00E+00 & 3.87E+02$=$ & 0.00E+00 & 3.87E+02$=$ & 0.00E+00 \\
$F_{26}$ & 1.01E+03 & 4.84E+01 & 1.03E+03$+$ & 5.60E+01 & 9.38E+02$-$ & 3.77E+01 & 9.33E+02$-$ & 3.92E+01 & 9.28E+02$-$ & 3.64E+01 & 1.18E+03$+$ & 1.43E+02 \\
$F_{27}$ & 4.90E+02 & 3.86E+00 & 5.01E+02$+$ & 4.57E+00 & 4.98E+02$+$ & 7.29E+00 & 4.79E+02$-$ & 6.51E+00 & 5.04E+02$+$ & 5.50E+00 & 5.03E+02$+$ & 7.56E+00 \\
$F_{28}$ & 3.13E+02 & 3.46E+01 & 3.28E+02$=$ & 4.57E+01 & 3.04E+02$=$ & 2.21E+01 & 3.02E+02$=$ & 1.60E+01 & 3.30E+02$+$ & 4.86E+01 & 3.42E+02$+$ & 5.91E+01 \\
$F_{29}$ & 4.31E+02 & 2.93E+01 & 4.42E+02$=$ & 3.14E+01 & 4.46E+02$+$ & 1.39E+01 & 4.14E+02$-$ & 2.72E+01 & 4.34E+02$+$ & 6.46E+00 & 4.82E+02$+$ & 3.84E+01 \\
$F_{30}$ & 2.02E+03 & 1.46E+01 & 2.02E+03$=$ & 1.14E+01 & 1.97E+03$-$ & 1.10E+02 & 1.04E+03$-$ & 3.21E+02 & 1.98E+03$-$ & 4.71E+01 & 2.35E+03$=$ & 1.36E+03 \\
\hline 
\textbf{W/T/L} & - & - & \textbf{17/12/0} & - & \textbf{9/15/5} & - & \textbf{9/10/10} & - & \textbf{17/8/4} & - & \textbf{19/9/1} & - \\
\hline
\textbf{Friedman Rank} & \textbf{2.36} & - & 4.02 & - & 2.98 & - & 2.64 & - & 3.53 & - & 5.47 & - \\
\hline
\end{tabular}

}
\end{table*}

\subsection{Experimental Setup and Parameter Settings}
The performance of the proposed OMCDE was evaluated on the CEC 2017 benchmark test suite \cite{wu2016problem}, which consists of 29 test functions ($F_1 - F_{29}$) comprising unimodal, multimodal, hybrid, and composition problems. All experiments were conducted on the Google Colab platform using an Intel Xeon CPU @ 2.20GHz with 12GB of RAM. The algorithm was implemented in Python 3.10 utilizing the NumPy library for vectorized matrix operations.

To ensure a fair comparison, the experimental settings were strictly maintained according to the CEC 2017 guidelines. For each function, 31 independent runs were executed with a dimensionality of $D=30$. The maximum number of function evaluations ($FES_{max}$) was set to $10,000 \cdot D$, and the search space bounds were fixed at $[-100, 100]^D$. The error value $f(\mathbf{x}_{best}) - f(\mathbf{x}^*)$ was recorded as the primary performance metric, where $f(\mathbf{x}^*)$ is the global optimum.

The control parameters are set as: population size $N_{init} = 18D$ linearly reduced to $N_{min} = 4$, archive size $|A| = 1.4N$, memory size $H = 6$ for $M_F$ and $M_{CR}$, and stage switching at $\lambda = 0.75$. Control parameters were adopted from established historical benchmarks. The stage-switching threshold $\lambda = 0.75$ was selected based on the Friedman mean ranking across all test functions to identify the optimal transition point

\subsection{Statistical Results and Analysis}
To evaluate the robustness of OMCDE on widely accepted historical benchmarks, we compared its performance on the CEC 2017 \cite{wu2016problem} test suite ($D=30$) against five established algorithms: TCDE \cite{yang2024triple}, LSHADE-RSP \cite{stanovov2018lshade}, iLSHADE-RSP \cite{choi2021improved}, L-SHADE \cite{tanabe2014improving}, and JADE \cite{zhang2009jade}. These competitors were selected to benchmark OMCDE against both its direct predecessor TCDE and the high-performing LSHADE family. Note that function $F_2$ is excluded from the analysis as it was officially removed from the CEC 2017 competition due to instability.

To rigorously validate these findings, we applied Friedman mean ranking test to provide a global performance assessment. As shown at the bottom of Table \ref{tab:cec2017}, OMCDE achieves the best overall rank of 2.36, followed by iLSHADE-RSP (2.64) and LSHADE-RSP (2.98), confirming its superior robustness across the entire benchmark suite.
Furthermore, Wilcoxon rank-sum test ($\alpha = 0.05$) was conducted, where symbols $-$, $+$, and $=$ indicate if a competitor is significantly better, worse, or equivalent to OMCDE. The analysis reveals a clear performance trajectory across landscape types. On Unimodal functions, OMCDE matches the perfect convergence of state-of-the-art methods, confirming its exploitation efficiency. In Simple Multimodal tasks, the Triplet Competition mechanism ensures sufficient diversity to escape local optima, consistently outperforming baselines like TCDE. Most importantly, OMCDE exhibits exceptional robustness on the challenging Hybrid and Composition functions ($F_{11}-F_{30}$). By effectively balancing early-stage exploration with late-stage greedy exploitation, the algorithm navigates these deceptive, non-separable landscapes more effectively than L-SHADE variants, securing significant wins on complex problems where traditional adaptive mechanisms often stagnate.

Table \ref{tab:cec2017} summarizes the comparative results over 31 independent runs. The results indicate that OMCDE exhibits superior performance across the test suite, particularly against its predecessor TCDE (17/12/0) and JADE (19/9/1), where it was only outperformed on a single function. It also maintains a significant lead over L-SHADE (17/8/4) and remains highly competitive against state-of-the-art variants LSHADE-RSP (9/15/5) and iLSHADE-RSP (9/10/10). While parity is maintained on unimodal landscapes, the algorithm demonstrates exceptional robustness on complex Hybrid ($F_{11}-F_{20}$) and Composition ($F_{21}-F_{30}$) functions. This success is primarily driven by the stage-dependent mutation, which enforces a more greedy scheme in final iterations to encourage superior exploitation around the global optimum. Complementing this, the Triplet Competition mechanism provides the necessary diversity to prevent the local optima stagnation common in traditional adaptive variants, securing a best-in-class Friedman rank of 2.36.

\subsection{Empirical Runtime Analysis}

\begin{figure}[htbp]
\centering
\includegraphics[width=0.9\linewidth]{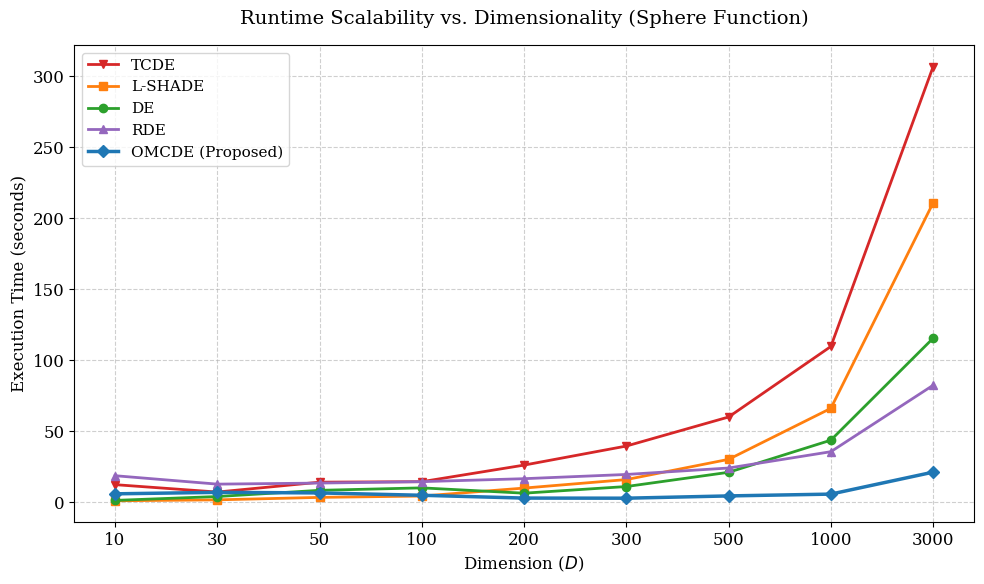}
\caption{Runtime scalability analysis comparing the vectorized OMCDE against a other variants of DE on the Sphere function ($N=100, FES=10^5$). The results highlight the superior efficiency of the matrix-based framework, which drastically reduces the computational overhead as dimensionality increases from $D=30$ to $D=3000$, unlike steep linear growth of the standard loop-based implementation.}\label{fig:runtime}
\end{figure}

To quantify the practical efficiency gains yielded by the matrix-based formulation, we conducted a runtime scalability experiment comparing the execution time of OMCDE against several standard Differential Evolution implementations. To isolate the algorithmic overhead from the objective function complexity, the computationally inexpensive Sphere function was selected as the test bed. All experiments were run with a population size of $N=100$ for $100,000$ Function Evaluations on varying dimensions $D \in \{10, 30, 50, 100, 200, 300, 500, 1000, 3000\}$ using the same computing environment.

Figure \ref{fig:runtime} illustrates the execution time as a function of dimensionality. As observed, while the runtime for all algorithms increases with dimension, the slope for OMCDE is significantly flatter than that of the traditional loop-based approaches. This demonstrates that as the dimensionality and problem scale increase, the vectorized matrix operations of OMCDE effectively reduce the computational overhead, resulting in a much faster and more scalable algorithm compared to its iterative counterparts.

\section{Conclusion}
In this study, we proposed the Opposition-based Matrix-Based Competitive Differential Evolution (OMCDE), a unified framework that integrates Opposition-Based Learning, Triplet Competition, and Stage-Dependent Mutation into fully vectorized matrix operations. By replacing traditional iterative logic with parallel linear algebra computations, OMCDE significantly enhances computational throughput while maintaining robust search capabilities. Our experiments on the CEC 2017 benchmark test suite demonstrate that this matrix-based formulation effectively balances exploration and exploitation, offering a competitive and highly efficient alternative to standard loop-based evolutionary algorithms. The success of OMCDE suggests that future differential evolution research can achieve significant efficiency gains by prioritizing vectorized implementations over complex control structures. 
Future work will focus on applying OMCDE to high-dimensional real-world optimization challenges and large-scale engineering design problems. Additionally, the potential of stage-dependent mutation strategies will be explored further. While the current mechanism imposes a global switch, future research should investigate adaptive frameworks where the transition between exploration and exploitation is determined at the individual level, driven by the unique evolutionary history and local landscape feedback of each candidate solution.

\section*{Acknowledgment}
The authors acknowledge the use of a Gemini LLMS  for refining the grammatical structure and stylistic presentation of the text. All algorithmic concepts, mathematical formulations, experimental designs, and scientific conclusions presented in this manuscript are the original work of the authors and were not generated by any LLMS.

\bibliographystyle{IEEEtran}
\begingroup
\renewcommand{\footnotesize}{\fontsize{7.2}{8.2}\selectfont}
\bibliography{main}
\endgroup

\end{document}